\documentclass[final,3p,times,twocolumn]{elsarticle}

\usepackage{amssymb}
\usepackage{amsmath}
\usepackage{comment}
\usepackage{float}
\usepackage{lipsum}
\usepackage{amsthm}
\usepackage{lineno}
\usepackage{hyperref}

\journal{SoftwareX}

\begin{document}

\begin{frontmatter}



\title{IntLevPy: A Python library to classify and model intermittent and L\'evy processes}

\author[oslomet,ailab]{Shailendra Bhandari}
\author[oslomet,ailab]{Pedro Lencastre}
\author[oslomet]{Sergiy Denysov}
\author[karkow]{Yurii Bystryk}
\author[oslomet,ailab,kuas,simula]{Pedro G. Lind}

\affiliation[oslomet]{organization={Department of Computer Science, OsloMet -- Oslo Metropolitan University},
             addressline={Pilestredet 52},
             city={Oslo},
             postcode={N-0130},
             country={Norway}}
\affiliation[ailab]{organization={AI Lab -- OsloMet Artificial Intelligence Lab},
             addressline={Pilestredet 52},
             city={Oslo},
             postcode={N-0166},
             country={Norway}}
\affiliation[kuas]{organization={School of Economics, Innovation and Technology, Kristiania University of Applied Sciences},
             addressline={Kirkegata 24-26},
             city={Oslo},
             postcode={N-0153},
             country={Norway}}
\affiliation[karkow]{organization={Institute of Applied Physics, National Academy of Sciences of Ukraine},
             addressline={ Petropavlivska Street 58},
             city={Sumy},
             postcode={40000},
             country={Ukraine}}
\affiliation[simula]{organization={Simula Research Laboratory, Numerical Analysis and Scientific Computing},
             addressline={},
             city={Oslo},
             postcode={N-0164},
             country={Norway}}
             
\begin{abstract}
IntLevPy provides a comprehensive description of the IntLevPy Package, a Python library designed for simulating and analyzing intermittent and L\'evy processes. The package includes functionalities for process simulation, including full parameter estimation and fitting optimization for both families of processes,
moment calculation, 
and classification methods. The classification methodology utilizes adjusted-$R^2$ and a noble performance measure $\Gamma$, enabling the distinction between intermittent and L\'evy processes. IntLevPy integrates iterative parameter optimization with simulation-based validation. This paper provides an in-depth user guide covering IntLevPy software architecture, installation, validation workflows, and usage examples. 
In this way, 
IntLevPy facilitates systematic exploration of these two broad classes of stochastic processes, bridging theoretical models and practical applications.

\end{abstract}

\begin{keyword}
{Intermittent processes \sep L\'evy flights \sep Stochastic simulations}



\end{keyword}

\end{frontmatter}

\begin{table*}
\centering
\begin{tabular}{|p{6cm}|p{9cm}|}
\hline
\textbf{Code metadata description} & \textbf{Metadata} \\
\hline
Current code version & \href{https://pypi.org/project/IntLevPy/}{v0.0.4}
 \\
\hline
Permanent link to code/repository used for this code version & \url{https://github.com/shailendrabhandari/IntLevPy.git} \\
\hline
Permanent link to Python Library & \url{https://pypi.org/project/IntLevPy/} \\
\hline
Legal Code License & MIT License \\
\hline
Code versioning system used & PyPI and git \\
\hline
Software code languages, tools, and services used & Python \\
\hline
Compilation requirements, operating environments \& dependencies & Requires Python 3.6 or later. Dependencies: numpy, scipy, matplotlib, pandas, seaborn, scikit-learn, pomegranate \\
\hline
If available, link to developer documentation/manual & \url{https://intlevpy.readthedocs.io/en/latest/} \\
\hline
Support email for questions & \href{mailto:shailendra.bhandari@oslomet.no}{shailendra.bhandari@oslomet.no} \\
\hline
\end{tabular}
\caption{Code metadata of the IntLevPy library.}
\label{codeMetadata} 
\end{table*}


\section{Motivation and significance}

In theoretical ecology and various other fields, understanding and modeling the movement patterns of organisms and particles is essential for unraveling complex systems dynamics. Random walks are fundamental tools used to simulate these stochastic movements observed in nature \cite{Patlak1953,Codling2008}. These models are vital for understanding how animals explore and exploit their environments to find food. Among the various models of random walks, Intermittent search (IS) and L\'evy walks (LW) 
strategies stand out \cite{Pyke2014}.

IS processes \cite{RevModPhys.83.81,Benichou_2006} are characterized by alternating between two distinct movement phases: ballistic and diffusive. During the ballistic phase, the searcher moves rapidly in a straight line, enabling it to cover large distances efficiently. In contrast, the diffusive phase involves slow, random movements, allowing for thorough local exploration. The dynamics of these phases can be mathematically described as:
\begin{equation}
    \vec{R}(t+\Delta t) = \vec{R}(t) + \left\{
    \begin{array}{ll}
        \vec{V}_{\rm B} \Delta t, & \text{(Ballistic)},  \\
        \vec{V}_{\rm D} \sqrt{\Delta t}, & \text{(Diffusive)},  
    \end{array}
    \right.
    \label{eq:intermittent}
\end{equation}
where $\vec{V}_{\rm B}$ is the constant velocity vector during the ballistic phase, and $\vec{V}_{\rm D} = D \vec{\eta}$ represents the random velocity during the diffusive phase, with $D$ being the diffusion coefficient and $\vec{\eta}$ a random vector with components drawn from a normal distribution with zero mean and unit variance. The transitions between phases are governed by the switching rates $\lambda_{BD}$ (from ballistic to diffusive) and $\lambda_{DB}$ (from diffusive to ballistic). The intermittent search model is thus defined by four key parameters: ${D, V_B, \lambda_{BD}, \lambda_{DB}}$.

LWs \cite{VISWANATHAN2008133,viswanathan_daluz_raposo_stanley_2011,RevModPhys.87.483} are a class of random walks known for their step lengths following a heavy-tailed power-law distribution, which leads to super-diffusive behavior. "Super"-diffusivity means they are characterized by frequent short steps interspersed with rare long ones. 
These two features are embedded in a power law  probability density function (PDF) for the step lengths $l$, namely 
\begin{equation}
    p(l) = \frac{\nu l_{\text{min}}^{\nu}}{l^{\nu + 1}},
\end{equation}
where 
$\quad l \geq l_{\text{min}}$,
$1 < \nu < 2$ is the L\'evy exponent, and $l_{\text{min}}$ is the minimum step length. The effectiveness of LW as an optimal search strategy was highlighted by Shlesinger and Klafter \cite{Shlesinger1986}, and further demonstrated by Viswanathan et al. \cite{Viswanathan1999}, who showed that an LW with $\ nu=1$ (Cauchy walk) optimizes search efficiency in sparse environments.
Typically we modelled LWs as uniform planar processes \cite{Rebenshtok2016}, 
where a particle moves at a constant velocity $v$ in a randomly chosen direction $\phi \in [0, 2\pi)$ for a time $\tau$, after which a new direction and time duration are selected based on the PDF:
\begin{equation}
\psi(\tau) = \frac{\gamma \tau_0^\gamma}{\left(\tau + \tau_0\right)^{1+\gamma}}, \label{eq:levy_pdf}
\end{equation}
with 
scale parameter $\tau_0 > 0$. The parameters ${\tau_0, \gamma, v}$ fully define the process.

Anomalous diffusion processes like IS and LW are ubiquitous in natural and artificial systems, including animal foraging patterns, microbial movement, human travel behavior, and even financial market dynamics \cite{Viswanathan1999,Bartumeus2002,Brockmann2006}. Accurate simulation and analysis of these processes are crucial for predicting system behavior and understanding underlying mechanisms. 
Recently, we have derived analytical expressions for the second and fourth-order moments of IS processes, as well as an accurate approximation of the moments of the LW class of processes. Moreover, we showed how these two statistical moments can be fitted with an iterative optimization scheme to estimate all parameters defining the underlying process.
Details are given in Ref.~\cite{lencastre2025}.
However, despite the importance of these processes, e.g.~to model time-series of measurements or observations,
there has been a lack of comprehensive, user-friendly computational libraries that enable researchers to simulate and analyze intermittent and L\'evy processes effectively. 
Existing tools often require extensive expertise to implement and may not offer the flexibility needed for diverse research applications. To overcome this shortcoming, we present IntLevPy.

The IntLevPy library is a Python library designed for simulating, analyzing, and classifying intermittent and L\'evy processes. 
Its main functionalities include (details below) the simulation of each class of processes, their statistical analysis (computation of moments), classification to distinguish between IS and LWs, and respective estimation of parameters.
With such features, the package contributes to scientific discovery by allowing researchers to explore new questions related to anomalous diffusion and search strategies in complex environments. For instance, it can be used to test hypotheses about optimal foraging strategies in ecology or to model the spread of information in social networks.

\section{Software description}

The IntLevPy is a modular Python library that provides tools for simulating and analyzing intermittent and L\'evy processes. The software is designed to be extensible and easy to integrate into existing workflows. Its code metadata descriptions are given in Table \ref{codeMetadata}.

\subsection{Software architecture}

The package is organized into the following modules:

\begin{itemize}
    \item \texttt{processes.py}: Functions for simulating intermittent and L\'evy processes.
    \item \texttt{moments.py}: Functions for calculating theoretical and empirical statistical moments.
    \item \texttt{optimization.py}: Functions for optimizing model parameters using empirical data.
    \item \texttt{classification.py}: Functions for classifying processes using statistical methods.
    \item \texttt{utils.py}: Utility functions for data analysis and processing.
    \item \texttt{examples/}: Directory containing example scripts demonstrating package usage.
    \item \texttt{tests/}: Directory containing unit tests for validating the code.
\end{itemize}

 \subsection{Software functionalities}
 
The major functionalities of the software include:

\begin{itemize}
\item \underline{Process simulation}: Generate realistic intermittent and L\'evy trajectories with customizable parameters, allowing for extensive experimentation and modeling.
\item \underline{Statistical analysis}: Calculate theoretical and empirical statistical moments (e.g., second and fourth moments). These statistical metrics are useful to understand the processes' dynamics and properties \cite{lencastre2025}.
\item \underline{Process classification}: Distinguish between intermittent and L\'evy processes based on statistical properties, aiding in analyzing experimental data and identifying underlying movement strategies.
\item \underline{Parameter estimation and optimization}: Fit model parameters to empirical data using advanced optimization techniques, enhancing model accuracy and reliability.
\end{itemize}

\begin{figure}[t]
\centering
\includegraphics[width=0.5\textwidth]{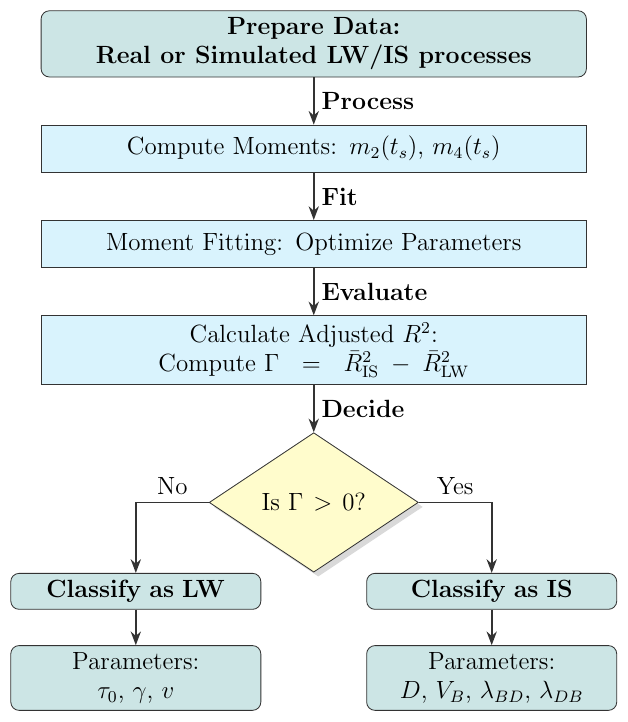}
\caption{Flowchart for classifying intermittent and L\'evy walk processes through iterative parameter optimization and statistical analysis. The methodology includes computation of moments, classification metrics, and iterative refinement of parameters.}
\label{fig:flowchart1}
\end{figure}

Figure \ref{fig:flowchart1} illustrates the detailed workflow of simulating, classifying, and modeling intermittent and L\'evy processes. Starting with parameter initialization and iterative random parameter generation, the workflow encompasses simulation, computation of the second and fourth moments, classification using statistical methods, and optimization of model parameters. The adjusted $R^2$ values drive the decision-making process, and calculating $\Gamma$ helps refine the classification between intermittent and L\'evy processes. The model then adapts accordingly to generate either an intermittent or a L\'evy process, highlighting the flexibility of the approach.


\section{Illustrative examples}



\begin{figure*}[t]
    \centering
    \includegraphics[width=\linewidth]{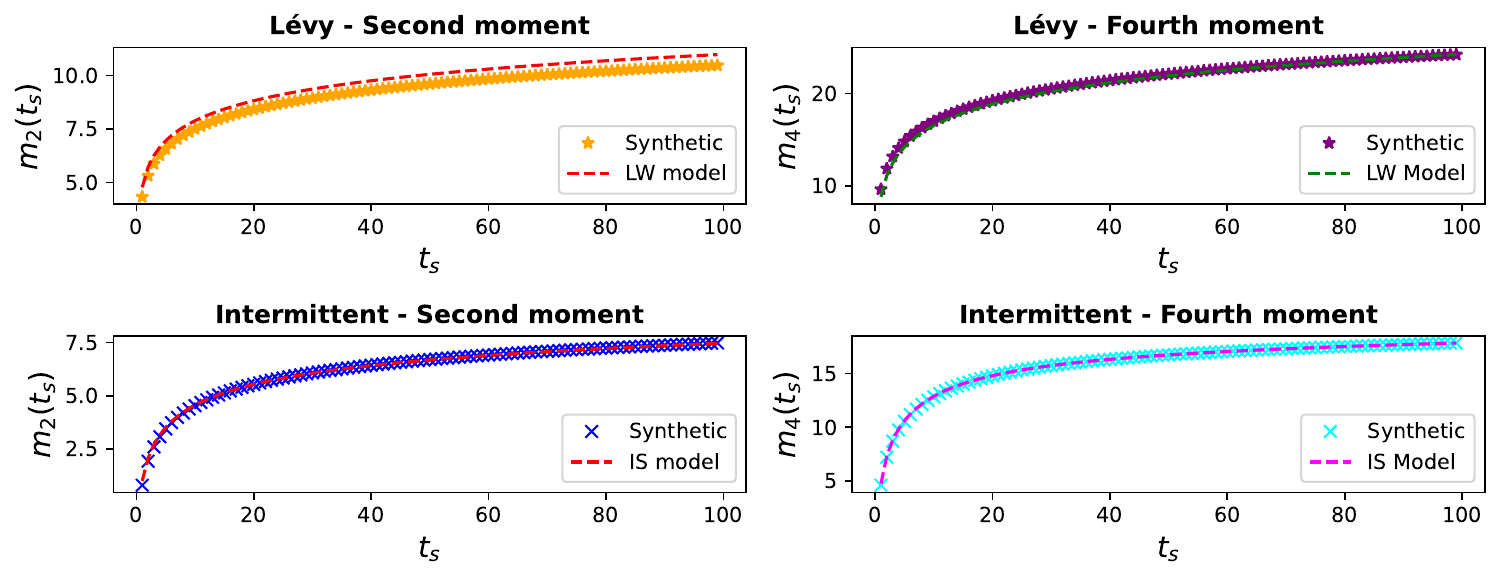}
    \includegraphics[width=\linewidth]{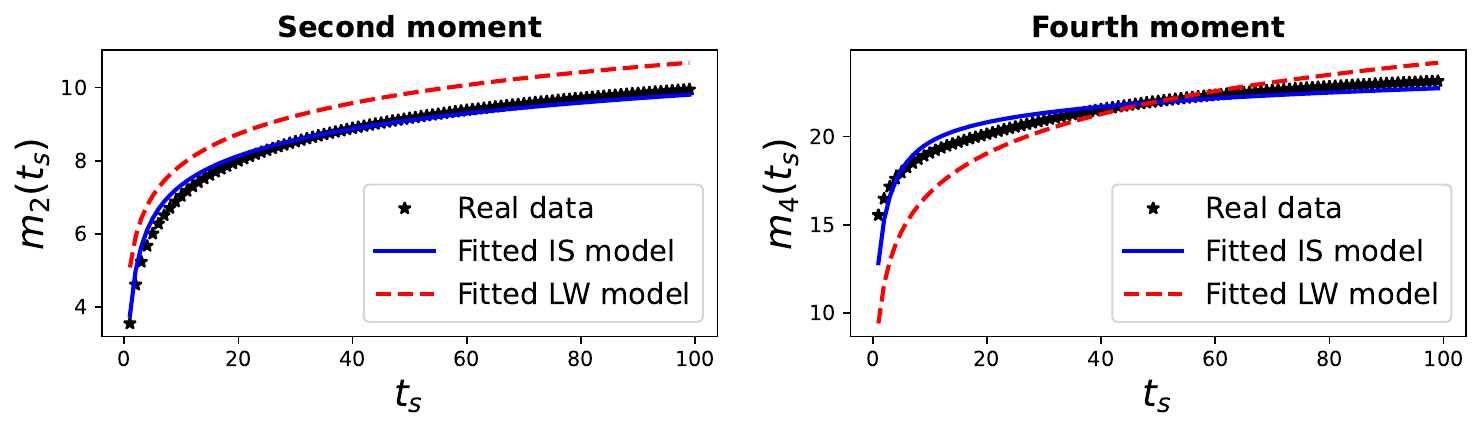}

    \caption{
    Comparison of synthetic, model-derived, and empirical moments for L\'evy and intermittent processes. 
    The top row shows the second moment (\(m_2(t_s)\)) and fourth moment (\(m_4(t_s)\)) for the L\'evy process (left) and intermittent process (right), where synthetic data is represented by markers and model fits are shown as dashed lines. 
    The bottom row compares empirical eye-tracking data with fitted models. The second moment (\(m_2(t_s)\)) and fourth moment (\(m_4(t_s)\)) are displayed as functions of the time scale \(t_s\). Empirical data is shown as black circles, the fitted intermittent search (IS) model as blue solid lines, and the fitted L\'evy walk (LW) model as red dashed lines.
    The close alignment of the empirical data with the IS model suggests a superior fit compared to the LW model, supporting the classification of the trajectory as an intermittent search process.
    }
    \label{fig:fitting}
\end{figure*}

Using IntLevPy, we analyzed real eye-tracking data to classify the underlying process followed by IS or LW. 

The process is characterized by a two-dimensional trajectory with points located at $(X_t,Y_t)$ where $t$ stamps the time-step.
To classify the observed trajectory, we begin by estimating the empirical $k$-order moments,
\begin{equation}
    m_k^{\text{exp}}(t_s) = \langle \| \vec{v}(t,t_s) \|^k\rangle_t \,,
\end{equation}
where the velocity components are defined as
\[v_X(t,t_s) = \frac{X_{t+t_s} - X_{t}}{t_s} \,  , 
\]
with similar expression for the $Y$-direction, and
\[\|\vec{v}(t,t_s)\| = \sqrt{v_X^2(t,t_s) + v_Y^2(t,t_s)}\, ,\]
which is the amplitude of the planar velocity.
Here we only consider second and fourth moments ($k=2,4$).
Having defined the moments of the process, we then introduce a quantity to evaluate how close an estimated moment is from the respective moment of experimental data:
\begin{equation}
    d_{k}(t_s)=\left [ \log
    \left (
    \frac{m^{\text{exp}}_k(t_s)}{m^{\text{data}}_k(t_s)} 
    \right ) \right ]^2 \, .
\end{equation}
This expression is used for each LW and IS independently.
Parameter optimization for each class of processes is carried out by minimizing $\overline{d}_2$ and $\overline{d}_4$ over the available dataset. After fitting the parameters, we compute adjusted-$R^2$-type measures for each model ($\overline{R}_{\text{IS}}^2$ and $\overline{R}_{\text{LW}}^2$). These measures characterize how well each model describes the empirical moments.

To classify the process, we define a score function
\begin{equation}
\Gamma = \overline{R}_{\text{IS}}^2 - \overline{R}_{\text{LW}}^2.
\end{equation}
If $\Gamma > 0$, the IS model yields a better fit, and we classify the underlying dynamics as an intermittent search process. Otherwise, if $\Gamma \leq 0$, the L\'evy walk model is favored.

Figure~\ref{fig:fitting} illustrates this procedure. The figure compares the empirically measured second and fourth moments $m_2^{\text{exp}}(t_s)$ and $m_4^{\text{exp}}(t_s)$ (black points) with those predicted by the optimized IS and LW models (colored curves). We observe that the IS model’s predictions align closely with the empirical data, especially for the fourth moment, resulting in higher $\overline{R}_{\text{IS}}^{2}$ values. Consequently, we find $\Gamma > 0$, indicating that the trajectory can be classified as stemming from an intermittent search process. We highlight its effectiveness and versatility by successfully applying IntLevPy to real data and demonstrating how it can distinguish between different classes of movement patterns. The ability to accurately classify underlying processes from observational data is crucial for understanding complex behaviors across a wide range of scientific domains.

\section{Impact}

The IntLevPy package significantly advances the study of anomalous diffusion processes by providing a comprehensive toolkit for simulating, analyzing, and classifying intermittent and L\'evy processes. By integrating simulation capabilities with advanced statistical analysis and optimization techniques, the package enables researchers to pursue new research questions regarding movement patterns and search strategies in complex systems \cite{Viswanathan1999,Bartumeus2005,Reynolds2018}. One of the primary impacts of IntLevPy is its ability to facilitate the classification of real-world trajectories, such as eye-tracking data, into IS or LW models. As illustrated in our analysis (see Figure \ref{fig:flowchart1}), the package implements a methodological framework that involves computing empirical moments from real data, estimating initial parameters, simulating both IS and LW models, optimizing parameters through iterative refinement, and ultimately classifying the trajectory based on statistical metrics like the adjusted R-squared and the score function $\Gamma$. In our illustrative examples, we applied IntLevPy to both synthetic and real data. For synthetic data, the package successfully recovered the known parameters of the simulated processes, demonstrating its robustness in parameter estimation and model fitting. When applied to real eye-tracking data, the package was able to classify trajectories effectively, distinguishing between intermittent and L\'evy behaviors based on the calculated moments and optimized model parameters. This capability opens new avenues for research in fields such as neuroscience, psychology, and ecology, where understanding the underlying mechanisms of movement patterns is crucial.

Furthermore, IntLevPy streamlines the research process by providing ready-to-use functions for complex tasks such as moment calculations and parameter optimization, which were previously time-consuming and error-prone when implemented from scratch. This efficiency allows researchers to focus on interpreting results and formulating new hypotheses rather than on the technical details of implementation. The potential applications of IntLevPy extend beyond academic research. The package can be utilized in commercial settings by enabling accurate modeling and prediction of complex diffusion processes, such as optimizing search and rescue operations, improving logistic networks, and enhancing financial market analyses \cite{Brockmann2006,Mantegna2000}. Its modular design and comprehensive documentation make it accessible to practitioners in various industries, potentially leading to innovations and improvements in operational efficiencies.


\section{Conclusions}
The IntLevPy package significantly contributes to the toolkit available for researchers and practitioners working with complex diffusion processes. By providing a unified framework for simulating intermittent and L\'evy processes 
or extract the best parameter choices for modelling a specific series of measurements,
the library addresses a critical need in the analysis of anomalous diffusion phenomena. 
Moreover, through the analytical expression of the statistical moments for IS and an accurate approximation of them for LWs, this library enables to distinguish accurately between both classes of processes.

All in all, our analysis demonstrates the efficacy of the package's methodological framework. The iterative optimization and classification process not only accurately fits models to empirical data but also provides insights into the underlying dynamics of the observed processes. By facilitating the distinction between IS and LW, IntLevPy enables a deeper understanding of movement patterns in various contexts, from animal foraging behaviors to human eye-tracking movements. This distinction is crucial for developing accurate models that can predict future behaviors and inform interventions or optimizations in practical applications. IntLevPy empowers researchers to conduct sophisticated analyses of complex diffusion processes with greater ease and accuracy. Its impact is expected to grow as it becomes an integral tool in the study of anomalous diffusion, contributing to advancements in both theoretical understanding and practical applications across multiple disciplines.
\section*{Acknowledgments}
The authors gratefully acknowledge support from the Research Council of Norway through the “VirtualEye” project (Ref.~335940-FORSKER22).


 \bibliographystyle{elsarticle-num-names} 
  \bibliography{references}

\end{document}